\documentclass[11pt,a4paper]{article}
\usepackage[utf8]{inputenc}
\usepackage[T1]{fontenc}
\usepackage{amsmath,amssymb,amsthm}
\usepackage{graphicx}
\usepackage[colorlinks=true,allcolors=blue]{hyperref}
\usepackage[numbers]{natbib}
\usepackage{geometry}
\usepackage{setspace}
\title{Can a Lightweight Automated AI Pipeline Solve Research-Level Mathematical Problems?}

\author{
  Lve Meng \thanks{University of Science and Technology of China. Work done during an internship at Zhongguancun Academy. Email: \texttt{ml130@mail.ustc.edu.cn}}
  \and
  Weilong Zhao \thanks{Université Paris Cité. Email: \texttt{weilong.zhao@etu.u-paris.fr}}
  \and
  Yanzhi Zhang \thanks{Zhongguancun Academy. Email: \texttt{s-zyz25@bza.edu.cn}}
  \and
  Haoxiang Guan\thanks{Zhongguancun Academy. Email: \texttt{s-ghx25@bza.edu.cn}}
  \and
  Jiyan He \thanks{Zhongguancun Academy. Email: \texttt{hejiyan@zgci.ac.cn}}
}

\date{\today}
\begin{document}
\maketitle
\begin{abstract}
\noindent
Large language models (LLMs) have recently achieved remarkable success in generating rigorous mathematical proofs, with ``AI for Math'' emerging as a vibrant field of research \citep{ju2026aimathematicsprogresschallenges}. While these models have mastered competition-level benchmarks like the International Mathematical Olympiad \citep{huang2025winninggoldimo2025, duan2025goldmedallevelolympiadgeometrysolving} and show promise in research applications through auto-formalization \citep{wang2025ariaagentretrievaliterative}, their deployment via lightweight, natural-language pipelines for research problems remains underexplored. In this work, we demonstrate that next-generation models (e.g., Gemini 3 Pro, GPT-5.2 Pro), when integrated into a streamlined automated pipeline optimized for \textbf{citation-based verification}, can solve sophisticated research-grade problems. We evaluate our pipeline on two novel datasets: (1) the \textbf{ICCM \cite{iccm2025_website} problem sets} (comparable to the S.-T. Yau College Student Mathematics Contest \cite{yau_contest_intro}) proposed by leading mathematicians \citep{shanghai2026mathchallenge}, and (2) the \textbf{``First Proof'' problem set} \citep{abouzaid2026proof}, consisting of previously unpublished research questions. Our pipeline generated candidate proofs for all problems in the first two ICCM sets and the ``First Proof'' set. The solutions for the first two ICCM sets and Problem 4 of the ``First Proof'' set have been fully verified by our team. All generated proofs have been submitted to the official organization, and our generated results are publicly available at \url{https://github.com/ml1301215/question_sets-test_results}. We have open-sourced the code and developed a user-friendly UI for this workflow, accessible at \url{https://github.com/ml1301215/research-math-assistant}.
\end{abstract}

\section{Introduction}
Automating advanced mathematical reasoning, a long-standing goal in AI, has accelerated markedly with the development of large language models (LLMs). Models have rapidly progressed from solving grade-school word problems to achieving medal-level performance in prestigious competitions \citep{chen2025seedprover15masteringundergraduatelevel, deepseek-math-v2}, such as the International Mathematical Olympiad (IMO) \citep{huang2025winninggoldimo2025,duan2025goldmedallevelolympiadgeometrysolving}. This progression demonstrates the dual role of AI in mathematics: it provides robust tools for mathematical research while simultaneously serving as a premier testbed for advancing general reasoning in AI systems \citep{ju2026aimathematicsprogresschallenges}. This raises a fundamental question: does success in solving curated contest problems translate to the ability to assist with genuine mathematical research?

Research-level mathematics is qualitatively different. As noted by Fields Medalist Shing-Tung Yau, while AI excels at high-dimensional computation, mathematicians remain essential for tackling deep, long-standing problems, and the future lies in collaboration \citep{shanghai2026mathchallenge}. The core of research often lies not in answering well-posed questions but in formulating them and developing new frameworks \citep{abouzaid2026proof}. Current benchmarks, primarily based on competition problems, fail to capture this reality. A critical step forward is the introduction of benchmarks such as the ``First Proof'' problem set \citep{abouzaid2026proof}, which consists of previously unpublished, research-originated questions. This addresses a key limitation: data contamination, where a model's performance is inflated by having seen similar problems during training. Because it is constructed from problems sourced directly from research, the design of the “First Proof” problem set guarantees that solving them requires novel reasoning, moving beyond mere pattern matching.

Meanwhile, the mathematical community has begun to issue direct challenges. Recently, leading mathematicians at the International Congress of Chinese Mathematicians (ICCM) \cite{iccm2025_website} publicly posed a set of sophisticated problems, explicitly targeting AI systems to probe the “boundaries of human knowledge” \citep{shanghai2026mathchallenge}. These parallel developments---the creation of research-level benchmarks and the public issuance of expert challenges---create a timely evaluation framework for assessing AI's true research potential.

Bridging the gap between contest performance and research utility remains non-trivial. Auto-formalization methods, which translate statements into verifiable code (e.g., Lean 4 \cite{moura2021lean4}), offer guaranteed correctness \citep{wang2025ariaagentretrievaliterative} but impose a high technical barrier, limiting accessibility for mathematicians. We argue that a complementary path is crucial: developing lightweight, natural-language pipelines that can generate reliable, human-readable proofs. 
In this work, we demonstrate that by integrating next-generation LLMs (e.g., Gemini 3 Pro, GPT-5.2 Pro) into a streamlined automated pipeline optimized with a citation-based verification mechanism, we can reliably solve sophisticated research-grade problems.

\section{Methodology}

\subsection{The Pipeline}
Given its outstanding performance and lightweight design, we adopted the automated pipeline architecture proposed in \citep{huang2025winninggoldimo2025}, originally developed for IMO-level problems. Given the increased complexity of our target tasks (Yau competition and research level), we introduced two key modifications:

\begin{enumerate}
    \item \textbf{Domain-Specific Prompt Optimization:} We refined prompts to handle higher-order abstract reasoning, moving beyond high-school olympiad strategies to incorporate undergraduate and graduate-level conceptual frameworks.
    
    \item \textbf{Citation-Augmented Verification:} A critical limitation of previous pipelines was the tendency to hallucinate theorems or formulas without sufficient context. Although such outputs might be mathematically correct, they often lack verifiability and readability, rendering the proofs effectively unverifiable. To address this, we enforced a strict constraint: the model must provide specific bibliographic references for non-trivial claims and explain the role of each cited source in the argument.
\end{enumerate}

\subsection{Validation of the Method}
To validate the citation-augmented approach, we tested the pipeline on representative exercises from Kashiwara's classic text \textit{Categories and Sheaves} \citep{kashiwara2006categories}. To probe the capability boundaries of the AI, several problems were selected by an independent researcher who had manually solved all the exercises in this book. The AI not only produced correct proofs but also correctly cited specific sections of the book. This significantly improved the interpretability of the output, allowing readers---even those less familiar with the text---to verify the logical chain.

\section{Experiments and Results}
We test our pipeline on two primary sources comprising extremely challenging mathematical problems.

\subsection{ICCM Problem Sets}
We utilized the three sets of problems proposed by the International Congress of Chinese Mathematicians (ICCM). Each set comprises six problems spanning several prominent mathematical domains, such as combinatorics, algebra, analysis, and differential equations.

\subsubsection{Sets 1 and 2}
These correspond to the difficulty of the S.-T. Yau College Student Mathematics Contest \cite{yau_contest_intro}.

\textbf{Result:} Our pipeline successfully solved 100\% of the problems in these two sets. The verification of the AI-generated solutions was initially conducted by our team, which includes members with a background in pure mathematics and a recipient of the contest's all-around medal in the S.-T. Yau College Student Mathematics Contest \cite{yau_contest_intro}. The final, verified proofs have been compiled into PDFs and submitted to the ICCM organization.

\subsubsection{Set 3}
This set contains open problems. Section 1 includes famous conjectures unsolved for decades; Section 2 includes open problems related to Calabi-Yau manifolds.

\textbf{Result:} The AI failed to solve Section 1 (as expected for open conjectures). Section 2 was attempted but remains unverified due to the lack of specialized domain experts in our team.

\subsection{The "First Proof" Problem Set}
We conducted a complete test using the “First Proof” problem set \citep{abouzaid2026proof}, which consists of ten previously unpublished research-level questions originating from mathematicians’ ongoing work. All questions were tested on February 9, 2026, prior to the release of the official answers on February 13, 2026.

\textbf{Result:} Our pipeline claimed to have produced correct solutions for all ten problems. Given the complexity of the generated proofs and time constraints for human verification, we prioritized a thorough verification of Problem 4. Considering the pipeline’s demonstrated tendency to acknowledge its limits on genuinely intractable tasks (e.g., the open conjectures in ICCM Set 3, Section 1), its confident solutions across the entire “First Proof” set suggest a high probability of success for the remaining unverified problems.

\section{Discussion}
Our results indicate that the combination of simple automated pipelines and state-of-the-art LLMs has crossed a significant threshold, demonstrating a tangible capacity for sophisticated mathematical reasoning on research-grade problems. However, this technical advancement reveals a shifting bottleneck and several practical hurdles that must be addressed to integrate AI as a practical assistant in daily mathematical research.

\subsection{The Verification Bottleneck}
The primary challenge has shifted from \textbf{proof generation} to \textbf{efficient verification}. While our pipeline produced candidate solutions for the entire “First Proof” set in minutes, the rigorous verification of a single problem (Problem 4) required hours. This asymmetry underscores an urgent need for more sophisticated AI-assisted verification tools—whether through advances in formal methods, interactive semi-formal interfaces, or explainable reasoning frameworks to keep pace with the accelerating speed of AI-generated mathematics.

\subsection{Practical Hurdles for Applications at Scale}
Beyond verification, our testing and development process revealed several practical challenges in applying AI systems to authentic mathematical research:

\begin{enumerate}
    \item \textbf{Usability and Accessibility Gap}: Many practicing mathematicians are unfamiliar with the prompting techniques and the effective use of advanced AI systems. Therefore, lowering the technical barrier and developing intuitive, powerful tools are crucial for widespread adoption within the mathematical community.

    \item \textbf{Long-Context Reasoning}: Genuine research often involves grappling with long, interconnected chains of reasoning and multiple related sub-problems. As problems deepen, the required context length and the need for coherent long-term memory pose challenges to current AI architectures, potentially leading to fragmented or inconsistent reasoning.

    \item \textbf{Understanding Implicit Knowledge}: Mathematical literature frequently contains implicit steps, assumed background knowledge, and notational shortcuts. When an AI lacks deep, specific domain understanding, it can fail to comprehend these “jumps,” as evidenced by the observation that training on large corpora of raw arXiv papers alone (e.g., in early iterations like DeepSeekMath-V1 \citep{shao2024deepseekmathpushinglimitsmathematical}) did not yield expected gains. This highlights that merely scaling data is insufficient. A promising direction may involve employing AI to process mathematical literature at scale, explicitly reconstructing intermediate steps and completing logical chains, and subsequently using this augmented corpus for reinforcement learning or fine-tuning training.
\end{enumerate}

\subsection{Outlook and Future Work}
Despite these challenges, the field of ``AI for Math'' shows immense promise. We believe that 2026 will be a pivotal year for application of AI in math research. The future is likely to be defined by collaborative synergy: AI systems will handle computationally intensive exploration, suggest novel patterns, and assist in the tedious verification of sub-steps, thereby freeing mathematicians to concentrate on high-level conceptualization and creative problem solving \citep{ju2026aimathematicsprogresschallenges}. To realize this vision, future efforts should focus on developing more intuitive interfaces, building models with stronger reasoning coherence, and fostering deeper understanding of mathematical literature.

\appendix
\section{Case Studies}
To illustrate our pipeline's capacity for handling combinatorial complexity, its proficiency in citing high-quality references, and its potential for application in mathematical research, we present three detailed case studies. All detailed solutions by AI discussed in these case studies are available online.\footnote{\url{https://github.com/ml1301215/question_sets-test_results}}

\subsection*{Case Study 1: Combinatorial Optimization}

The first problem, drawn from Question 1 of ICCM’s initial problem set, presents a complex ranking and elimination scenario. It demands rigorous set-theoretic reasoning alongside constructive existence proofs. Problems of this combinatorial nature have traditionally been regarded as a challenge for artificial intelligence.

\textbf{Problem Statement:}
There are 8 students and 3 subjects. Students have distinct scores in each subject. A student is called a \textit{potential champion} if, in at least one of the six possible orderings of the subjects, they survive to the end, where each subject eliminates the bottom half of the remaining students. Find the maximum possible number of potential champions.

\textbf{AI Solution Overview:}
The AI successfully identified the maximum number as 5 and provided a rigorous proof. The solution strategy generated by the pipeline combined abstract set theory with targeted construction.

\textbf{Logic of the Proof:}
Let $U = \{1, \dots, 8\}$ be the set of students. For any subject $X$, let $S_X$ denote the set of the top 4 students in that subject. The AI derived the following key lemmas:

\begin{enumerate}
    \item \textbf{Intersection Lower Bound:} Using the Principle of Inclusion-Exclusion, the AI showed that $\sum_{X<Y} |S_X \cap S_Y| \ge 4$.
    \item \textbf{Intersection Size Constraint (Lemma 2):} The AI proved that if $|S_X \cap S_Y| = 2$, then the winner of the ordering $(X, Y, Z)$ is identical to the winner of $(Y, X, Z)$. This forces a reduction in the number of unique potential champions.
    \item \textbf{Impossibility of 6 Champions:} The AI analyzed the configuration of intersection sizes. To avoid the duplication caused by size-2 intersections, the configuration must rely on intersections of size 3 or 4. However, the AI rigorously demonstrated that:
    \begin{itemize}
        \item If $|S_A \cap S_B| = 3$, at least two winners among $\{w_{ABC}, w_{BAC}, w_{BCA}\}$ are identical.
        \item If $|S_A \cap S_B| = 4$ (i.e., $S_A = S_B$), the set of winners generated by permutations of $A, B, C$ contains at most 3 distinct students.
    \end{itemize}
\end{enumerate}
Concluding that 6 distinct champions are impossible, the AI constructed a concrete counterexample yielding exactly 5 champions:
\[
S_A = \{1, 2, 3, 5\}, \quad S_B = \{3, 4, 7, 8\}, \quad S_C = \{1, 2, 4, 6\},
\]
with specific ranks assigned to satisfy the elimination criteria, verifying the answer $N=5$.
Following the AI's successful proof, we attempted to formalize it, resulting in over 5,000 lines of formalized code in Lean 4.

\subsection*{Case Study 2: Category Theory (Exercise from Kashiwara \& Schapira \citep{kashiwara2006categories})}
The second case study evaluates the pipeline's capacity to handle abstract definitions and theorem proving within the context of category theory, as well as its proficiency in managing bibliographic citations.

\textbf{Problem Statement (Exercise 3.5 in \citep{kashiwara2006categories}):}
Let $F: \mathcal{C} \to \mathcal{C}'$ be a functor of small categories. Prove that $F$ is \textit{left exact} (meaning $\mathcal{C} \times_{\mathcal{C}'} \mathcal{C}'_{/Y}$ is cofiltered for every $Y$) if and only if the Yoneda extension $\widehat{F}: \mathcal{C}^{\wedge} \to \mathcal{C}'^{\wedge}$ is exact.

\textbf{Logic of the Proof:}
The pipeline correctly interpreted the definitions and established the equivalence chain. Notably, it correctly identified the ambiguity in standard terminology regarding "left exact functors" and anchored its proof to the definition used in the specific textbook cited.

\textbf{Main Steps in the AI Proof:}
\begin{enumerate}
    \item \textbf{Yoneda Extension:} The AI defined $\widehat{F}$ as the left Kan extension of $\mathrm{h}_{\mathcal{C}'} \circ F$ along $\mathrm{h}_{\mathcal{C}}$, noting that $\widehat{F}$ is a left adjoint and thus always right exact.
    \item \textbf{Reduction to Finite Limits:} Since $\widehat{F}$ is right exact, it is exact if and only if it is left exact (preserves finite limits).
    \item \textbf{Pointwise Computation:} The AI utilized the coend formula to show that $\widehat{F}(P)(Y') \cong \varinjlim_{(Y' \downarrow F)^{\mathrm{op}}} P$. Since limits in presheaf categories are computed pointwise, the exactness of $\widehat{F}$ depends on the exactness of the colimit functor in $\mathbf{Set}$.
    \item \textbf{Filtered Colimits:} The AI applied the standard theorem valid in $\mathbf{Set}$, which states that a colimit functor preserves finite limits if and only if the indexing category is \textit{filtered}.
    \item \textbf{Connecting Concepts:} Finally, the AI correctly linked the filteredness of $(Y' \downarrow F)^{\mathrm{op}}$ to the cofilteredness of the comma category $(Y' \downarrow F)$. It verified that this matches the textbook's definition of $F$ being left exact.
\end{enumerate}

The solution included precise references to Definition 3.3.18 in \cite{kashiwara2006categories} and correctly utilized relevant nLab entries to support intermediate steps. 

\subsection*{Case Study 3: Analytic Theory of Polynomials (Problem 4 from ``First Proof'' Set)}

The third case, selected from the ``First Proof'' problem set \citep{abouzaid2026proof}, illustrates the pipeline's capability to test the validity of a research-level problem by identifying counterexamples through rigorous analytical derivation.

\textbf{Problem Statement:}
Let $p(x)$ and $q(x)$ be two monic polynomials of degree $n$:
\[
p(x) = \sum_{k=0}^{n} a_k x^{n-k} \quad \text{and} \quad q(x) = \sum_{k=0}^{n} b_k x^{n-k},
\]
where $a_0 = b_0 = 1$. Define the operation $p \boxplus_n q$ as the polynomial
\[
(p \boxplus_n q)(x) = \sum_{k=0}^{n} c_k x^{n-k}, \quad \text{where } c_k = \sum_{i+j=k} \frac{(n-i)!(n-j)!}{n!(n-k)!} a_i b_j.
\]
For a monic polynomial $p(x) = \prod_{i \leq n} (x - \lambda_i)$, define the functional
\[
\Phi_n(p) := \left( \sum_{i \leq n} \prod_{j \neq i} \frac{1}{\lambda_i - \lambda_j} \right)^2,
\]
setting $\Phi_n(p) := \infty$ if $p$ has a multiple root. Determine whether the following inequality holds for all monic real-rooted polynomials $p(x)$ and $q(x)$ of degree $n$:
\[
\frac{1}{\Phi_n(p \boxplus_n q)} \geq \frac{1}{\Phi_n(p)} + \frac{1}{\Phi_n(q)}.
\]

\textbf{AI Solution Overview:}
The AI successfully identified that the statement is \textbf{false}. By analyzing the asymptotic behavior of the rational function $1/p(z)$, the AI derived an explicit formula for $\Phi_n(p)$ and constructed a definitive counterexample in the case $n=1$.

\textbf{Logic of the Proof:}
The AI's reasoning proceeded in three main steps:

\begin{enumerate}
    \item \textbf{Residue Analysis:} The AI identified the inner sum $S_n(p) = \sum_{i=1}^n \prod_{j \neq i} \frac{1}{\lambda_i - \lambda_j}$ as the sum of the residues of the function $f(z) = \frac{1}{p(z)}$ with distinct $\lambda_i$.
    
    \item \textbf{Laurent Expansion:} By expanding $1/p(z)$ as a Laurent series at infinity:
    \[
    \frac{1}{p(z)} = \frac{1}{z^n} \left( 1 + \frac{a_1}{z} + \dots \right)^{-1} = \frac{1}{z^n} - \frac{a_1}{z^{n+1}} + O\left(\frac{1}{z^{n+2}}\right),
    \]
    the AI compared coefficients to find $S_n(p)$.
    \begin{itemize}
        \item For $n=1$, the term $1/z$ has coefficient 1, so $S_1(p) = 1$.
        \item For $n \ge 2$, the expansion starts at $1/z^n$, meaning the coefficient of $1/z$ is 0. Thus, $S_n(p) = 0$ for $n \ge 2$.
    \end{itemize}
    Consequently, $\Phi_1(p) = 1$ and $\Phi_n(p) = 0$ for $n \ge 2$ (for distinct roots).

    \item \textbf{Counterexample Construction ($n=1$):} 
    The AI explicitly computed the operation for linear polynomials $p(x) = x + a_1$ and $q(x) = x + b_1$. The formula for coefficients yields:
    \[
    c_0 = 1, \quad c_1 = a_1 + b_1.
    \]
    Thus, $p \boxplus_1 q = x + (a_1 + b_1)$, which is also a monic polynomial of degree 1.
    Substituting these values into the proposed inequality:
    \[
    \frac{1}{1} \geq \frac{1}{1} + \frac{1}{1} \implies 1 \geq 2,
    \]
    which is a contradiction.
\end{enumerate}

The AI concluded that the inequality fails for $n=1$, thereby disproving the universal claim. This case demonstrates the pipeline's ability to efficiently simplify complex definitions into testable base cases. Moreover, since the problem arose from an authentic mathematical research process \cite{abouzaid2026proof}, the AI's success also highlights its practical potential in mathematical research.


\bibliographystyle{plainnat}
\bibliography{references}

@misc{abouzaid2026proof,
      title={First Proof}, 
      author={Mohammed Abouzaid and Andrew J. Blumberg and Martin Hairer and Joe Kileel and Tamara G. Kolda and Paul D. Nelson and Daniel Spielman and Nikhil Srivastava and Rachel Ward and Shmuel Weinberger and Lauren Williams},
      year={2026},
      eprint={2602.05192},
      archivePrefix={arXiv},
      primaryClass={cs.AI},
      url={https://arxiv.org/abs/2602.05192}, 
}

@misc{chen2025seedprover15masteringundergraduatelevel,
      title={Seed-Prover 1.5: Mastering Undergraduate-Level Theorem Proving via Learning from Experience}, 
      author={Jiangjie Chen and Wenxiang Chen and Jiacheng Du and Jinyi Hu and Zhicheng Jiang and Allan Jie and Xiaoran Jin and Xing Jin and Chenggang Li and Wenlei Shi and Zhihong Wang and Mingxuan Wang and Chenrui Wei and Shufa Wei and Huajian Xin and Fan Yang and Weihao Gao and Zheng Yuan and Tianyang Zhan and Zeyu Zheng and Tianxi Zhou and Thomas Hanwen Zhu},
      year={2025},
      eprint={2512.17260},
      archivePrefix={arXiv},
      primaryClass={cs.CL},
      url={https://arxiv.org/abs/2512.17260}, 
}

@misc{huang2025winninggoldimo2025,
      title={Winning Gold at IMO 2025 with a Model-Agnostic Verification-and-Refinement Pipeline}, 
      author={Yichen Huang and Lin F. Yang},
      year={2025},
      eprint={2507.15855},
      archivePrefix={arXiv},
      primaryClass={cs.AI},
      url={https://arxiv.org/abs/2507.15855}, 
}

@misc{wang2025ariaagentretrievaliterative,
      title={Aria: An Agent For Retrieval and Iterative Auto-Formalization via Dependency Graph}, 
      author={Hanyu Wang and Ruohan Xie and Yutong Wang and Guoxiong Gao and Xintao Yu and Bin Dong},
      year={2025},
      eprint={2510.04520},
      archivePrefix={arXiv},
      primaryClass={cs.AI},
      url={https://arxiv.org/abs/2510.04520}, 
}

@misc{ju2026aimathematicsprogresschallenges,
      title={AI for Mathematics: Progress, Challenges, and Prospects}, 
      author={Haocheng Ju and Bin Dong},
      year={2026},
      eprint={2601.13209},
      archivePrefix={arXiv},
      primaryClass={math.HO},
      url={https://arxiv.org/abs/2601.13209}, 
}

@misc{deepseek-math-v2,
  author = {Zhihong Shao and Yuxiang Luo and Chengda Lu and Z.Z. Ren and Jiewen Hu and Tian Ye and Zhibin Gou and Shirong Ma and Xiaokang Zhang},
  title = {DeepSeekMath-V2: Towards Self-Verifiable Mathematical Reasoning},
  year = {2025},
}

@misc{shanghai2026mathchallenge,
  title         = {Shanghai Issues the "Mathematical Questions" Again After Six Months: Three Challenging Math Problems Pushing AI to Its Limits},
  author        = {{Shanghai Institute of Mathematical Sciences and Interdisciplinary Sciences}},
  year          = {2026},
  month         = {1},
  day           = {9},
  howpublished  = {DongJing (WeChat Official Account Platform)},
  note          = {A non-peer-reviewed news report. It covers the event where Fields Medalist Shing-Tung Yau and other mathematicians publicly issued challenging math problems to AI systems. The article also mentions the performance of several AI models, including InternLM, ByteDance Seed, Qwen, and SenseTime, in tackling these problems. Suitable for citation as background information or to reference a current development.},
  url           = {https://mp.weixin.qq.com/s/v2KA8Cdoe-Nj0599GIfQMw},
  urldate       = {2026-02-11}
}

@book{kashiwara2006categories,
  title      = {Categories and Sheaves},
  author     = {Kashiwara, Masaki and Schapira, Pierre},
  series     = {Grundlehren der mathematischen Wissenschaften},
  volume     = {332},
  publisher  = {Springer},
  address    = {Berlin, Heidelberg},
  year       = {2006},
  isbn       = {978-3-540-27949-5},
  doi        = {10.1007/3-540-27950-4}
}

@misc{shao2024deepseekmathpushinglimitsmathematical,
      title={DeepSeekMath: Pushing the Limits of Mathematical Reasoning in Open Language Models}, 
      author={Zhihong Shao and Peiyi Wang and Qihao Zhu and Runxin Xu and Junxiao Song and Xiao Bi and Haowei Zhang and Mingchuan Zhang and Y. K. Li and Y. Wu and Daya Guo},
      year={2024},
      eprint={2402.03300},
      archivePrefix={arXiv},
      primaryClass={cs.CL},
      url={https://arxiv.org/abs/2402.03300}, 
}

@online{yau_contest_intro,
  author = {{S.-T. Yau College Student Mathematics Contest Organizing Committee}},
  title = {Introduction to S.-T. Yau College Student Mathematics Contest},
  url = {http://yau-contest.com/page-Introduction.html},
  urldate = {2026-02-13}
}

@misc{iccm2025_website,
  author       = {{International Congress of Chinese Mathematicians}},
  title        = {{The 10th International Congress of Chinese Mathematicians (ICCM 2025)}},
  howpublished = {\url{https://iccm.simis.cn/iccm2025/site/}},
  year         = {2025},
  note         = {Accessed: 2026-02-13}
}

@inproceedings{moura2021lean4,
  author    = {Leonardo de Moura and Sebastian Ullrich},
  title     = {{The Lean 4 Theorem Prover and Programming Language}},
  booktitle = {Automated Deduction -- CADE 28},
  year      = {2021},
  editor    = {Andr{\'e} Platzer and Geoff Sutcliffe},
  series    = {Lecture Notes in Computer Science},
  volume    = {12699},
  pages     = {625--635},
  publisher = {Springer},
  doi       = {10.1007/978-3-030-79876-5_37},
  url       = {https://doi.org/10.1007/978-3-030-79876-5_37}
}

@misc{duan2025goldmedallevelolympiadgeometrysolving,
      title={Gold-Medal-Level Olympiad Geometry Solving with Efficient Heuristic Auxiliary Constructions}, 
      author={Boyan Duan and Xiao Liang and Shuai Lu and Yaoxiang Wang and Yelong Shen and Kai-Wei Chang and Ying Nian Wu and Mao Yang and Weizhu Chen and Yeyun Gong},
      year={2025},
      eprint={2512.00097},
      archivePrefix={arXiv},
      primaryClass={cs.AI},
      url={https://arxiv.org/abs/2512.00097}, 
}

\end{document}